\documentclass{article} 
\usepackage{graphicx}
\usepackage{iclr2021_conference,times}
\iclrfinalcopy

\usepackage{amsmath,amsfonts,bm}

\def\eqref#1{equation~\ref{#1}}

\def\1{\bm{1}}

\DeclareMathAlphabet{\mathsfit}{\encodingdefault}{\sfdefault}{m}{sl}
\SetMathAlphabet{\mathsfit}{bold}{\encodingdefault}{\sfdefault}{bx}{n}

\usepackage{hyperref}
\usepackage{url}

\title{Heterogeneous network-based Drug Repurposing for COVID-19}

\author{Shuting Jin\textsuperscript{1,2}, Xiangxiang Zeng\textsuperscript{3}, Wei Huang\textsuperscript{1}, Feng Xia\textsuperscript{1}, Changzhi Jiang\textsuperscript{1}, Xiangrong Liu\textsuperscript{1} \& Shaoliang Peng\textsuperscript{3}\\
\textsuperscript{1}Department of Computer Science, Xiamen University, Xiamen 361005, China\\
\textsuperscript{2}National Institute for Data Science in Health and Medicine, Xiamen University, Xiamen 361005, China\\
\textsuperscript{3}School of Information Science and Engineering, Hunan University, Changsha 410082, China \\
\texttt{\{stjin,huangwii,xiafeng,czjiang\}@stu.xmu.edu.cn} \\
\texttt{\{xzeng,slpeng\}@hnu.edu.cn}\\
\texttt{xrliu@xmu.edu.cn}
}

\begin{document}

\maketitle

\begin{abstract}
The Corona Virus Disease 2019 (COVID-19) belongs to human coronaviruses (HCoVs), which  spreads rapidly around the world. Compared with new  drug development, drug repurposing may be the best  shortcut for treating COVID-19. Therefore, we constructed a comprehensive heterogeneous network based on the HCoVs-related target proteins and use the  previously proposed deepDTnet, to discover potential drug candidates for COVID-19. We obtain high performance in predicting the possible drugs effective for COVID-19 related proteins. In summary, this work utilizes a powerful heterogeneous network-based deep learning method, which may be beneficial to quickly identify candidate repurposable drugs toward future clinical trials for COVID-19. The code and data are available at https://github.com/stjin-XMU/HnDR-COVID.
\end{abstract}

\section{Introduction}

Coronavirus (CoVs) is a type of enveloped single-stranded RNA virus and the largest RNA virus known in the genome. HCoVs including the Severe Acute Respiratory Syndrome Coro-navirus (SARS-CoV), which caused  severe losses in 2003, and Middle East Respiratory Syndrome Coronavirus (MERS-CoV), which swept the Middle East in 2012, are coronaviruses with high morbidity and mortality worldwide \cite{paules2020coronavirus}. In 2019, a third pathogenic HCoV was named COVID-19, which is also a kind of coronavirus. By June 20, 2021, the outbreak of COVID-19 has caused more than one hundred million people infected and two million people died. Organizations around the world are working hard to develop prevention and treatment for COVID-19, but there are no effective  drugs. There is an urgent need to develop effective prevention and treatment strategies for the COVID-19 outbreak to prevent the virus from spreading further.\\
Most approved drugs do not directly target disease-related proteins, but bind proteins near their networks \cite{gysi2020network}. Our goal is to identify drug candidates that may directly target HCoVs-related proteins or may disrupt the coronavirus disease module network by targeting the neighbors of HCoVs-related proteins. In this study, based on the HCoVs-related target proteins proposed by existing studies, we construct a heterogeneous network connecting drugs, diseases, and HCoVs-related proteins, and apply a deep learning algorithm proposed by previous work \cite{zeng2020target} to identify drugs that may be related to HCoVs.

\section{MATERIALS AND METHODS}
\label{gen_inst}

\subsection{Methods}
The method framework is shown in Figure.1. Here, we utilize heterogeneous network based on deep learning method for discovering HCoVs protein-related drugs. Specifically, 1) Based on the target proteins related to the HCoVs proposed in the Zhou et al. \cite{zhou2020network}, we retain these proteins and the directly related proteins map to the PPI network for calculation. Based on these proteins, we integrate 17 networks including drug-protein, drug-disease, drug-drug, protein-disease, protein-protein associations network from existing related databases. 2) There are two main types of networks in this  experiment: homogeneous interaction networks and heterogeneous interaction networks. For heterogeneous interaction networks, we use the Jaccard similarity coefficient to calculate the similarity network. 3) We apply deepDTnet \cite{zeng2020target} algorithm to infer whether two drugs share a target and  predict novel drug-protein interactions, and find possible drugs that are effective for COVID-19 related proteins (see Methods).
\subsubsection{Constructing network Materials}
The materials of these networks are summarized in Table 1. Among them, drug-drug, drug-disease and 8  drug similarity networks directly use the data of deepDTnet work. For the rest of the target protein-related data, based on the HCoVs-related proteins \cite{zhou2020network}, we mapped  these proteins to the directly related proteins in the PPIs network, and retained these proteins to construct protein related networks.

\begin{table}[t]
\caption{Materials of networks}
\label{sample-table}
\begin{center}
\begin{tabular}{llll}
\multicolumn{1}{c}{\bf  }  &\multicolumn{1}{c}{\bf Node1} &\multicolumn{1}{c}{\bf Node2} &\multicolumn{1}{c}{\bf Edges}
\\ \hline \\

  Protein-protein & 5390 & 5390 & 136763 \\
  Protein-disease & 5390 & 3087 & 85018 \\
  Drug-protein & 2953 & 5390 & 5278 \\
  Drug-drug & 2953 & 2953 & 220736 \\
  Drug-disease & 2953 & 3087 & 6677 \\
  \multicolumn{4}{c}{4 protein similarity networks} \\
  \multicolumn{4}{c}{8 drug similarity networks} \\
\end{tabular}
\end{center}
\end{table}

\begin{figure*}
\begin{center}
\includegraphics[width=0.9\textwidth]{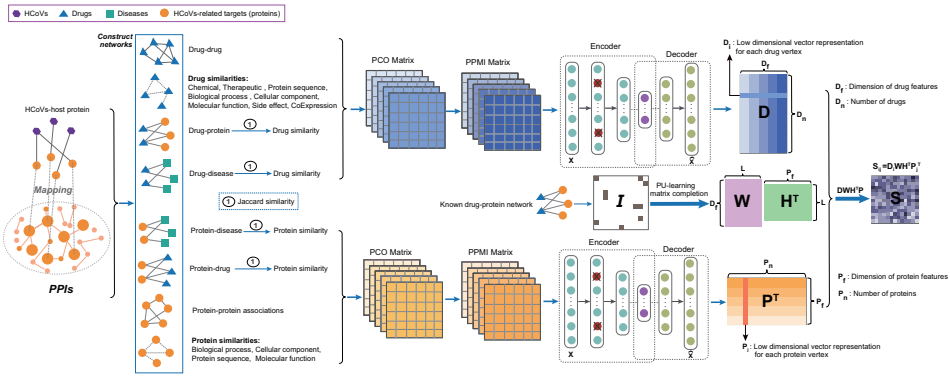}
\caption{A diagram illustrates the pipeline of the entire method. The method embeds the 17 types of chemical, genomic, phenotypic, and cellular networks and applies deepDTnet to predict identify candidate drugs toward COVID-19.}
\label{Figure.1}
\end{center}
\end{figure*}

\subsubsection{Construction of similarity networks}
 The Jaccard similarity coefficient is a very popular statistical method for calculating the similarity between two sets. Suppose we have two sets $A$ and $B$. The Jaccard similarity \cite{niwattanakul2013using} of these two sets is:

\begin{equation*}
Sim(A,B) = \frac{|A \cap B|}{|A \cup B|}
\tag{1}
\end{equation*}

\subsubsection{Drug-protein associations prediction}
For a certain network, the random surfing model is used to extract network co-occurrence probability information and generate a corresponding probabilistic co-occurrence (PCO) matrix.After yielding the PCO matrix, we calculate a shifted PPMI matrix inspired by \cite{bullinaria2007extracting}.\\
Then, we use SDAE, an advanced dimensionality reduction technology, which can help us to capture important topology information in the network and obtain better low-dimensional vectors through the learning of the PPMI matrix. The optimization process of SDAE is as follows:
\begin{equation*}
\min_{\{W_l\},\{b_l\}}||x - \hat{x}||_F^2 + \lambda \sum_t||W_l||_F^2
\tag{2}
\end{equation*}

Where $l$ is the number of layers, $W_{l}$ is the weight of the $l$ layer, and $b_{l}$ is a biased vector of the $l$ layer. $\lambda$ is a regularization parameter and $||\cdot||_{F}$ denotes the Frobenius norm. The entire SDAE has a total of $L$ layers, of which the first half is the encoder and the second half is the decoder, as shown in Fig.1.

Finally, we use PU learning for matrix completion \cite{hsieh2015pu}. In detail, we make the drug-protein interaction network $I$. The observed drug-protein interaction label is 1, while the unobserved is 0. After obtaining the characteristics $W$ and $H$ of the drug and target in the previous process, we use $I$ to perform matrix completion on $WH^{T}$. The specific optimization function is as follows:

\begin{equation*}
\begin{split}
&\min_{W,H}\sum(I_{ij} - D_iWH^TP_j^T)^2 + \\
&\alpha\sum_{(i,j)\in\Omega^-} (I_{ij} - D_iWH^TP_j^T)^2 + \lambda(|| W ||_F^2 + || H ||_F^2) \\
\end{split}
\tag{3}
\end{equation*}

Where $\Omega^+$ denotes the known response in the sample and $\Omega^-$  denotes the missing term selected as the negative samples. Finally, we can approximate the probability of drug and target response:

\begin{equation*}
Score(i, j) = D_iWH^TP_j^T
\tag{4}
\end{equation*}

\subsection{Verification Based on Heterogeneous Network Results}

To evaluate the performance, we firstly perform 5-fold cross-validation on the drug-protein prediction results. To reduce the data bias of  cross-validation, it was repeated 10 times and the average performance was computed. The AUROC value was  approximately 0.941 (Figure.2(a)) and the AUPR value was  approximately 0.954 (Figure.2(b)). It proves the feasibility  of the algorithm.\\
Because our work is mainly to discover drugs related to COVID-19 proteins, we verified the predicted  drug-protein of the 119 HCoVs-host proteins summarized in the research by \cite{zhou2020network}. We screen out the  results of the drug and protein that predicted by the 119 HCoVs-host proteins and rank the correlation of 2953 approved drugs in the data based on the predicted scores. Then the 45 drugs from COVID-19 related therapeutic in clinical trials were used as positive samples to verify the external data set (see Supplementary data Table 1). The obtained AUROC value was about 0.785 (Figure.2(a)) and the obtained AUPR value was about 0.803 (Figure.2(b)).

\begin{figure}[!htbp]
\centering
\includegraphics[width=0.7\textwidth]{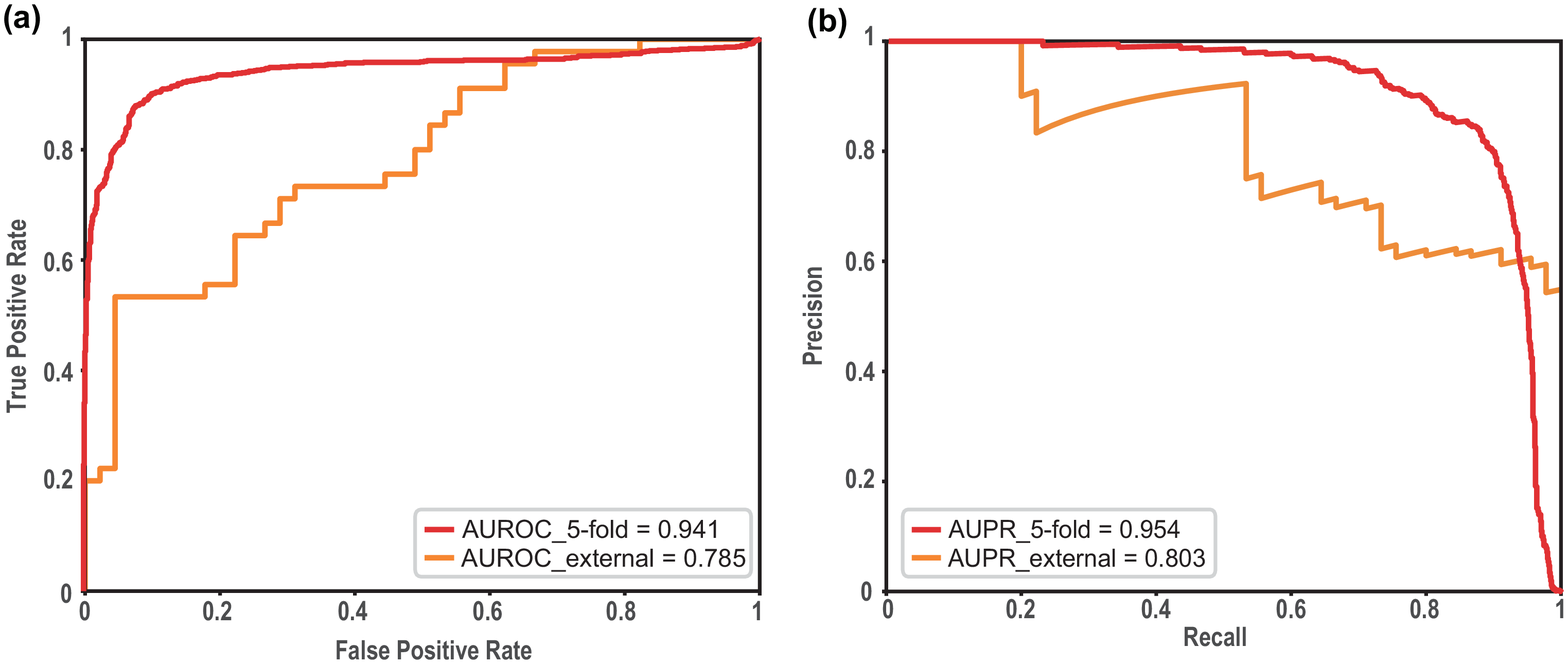}
\caption{Performance of the method was assessed by 5-fold cross-validation and the external validation set.}
\label{Figure.2}
\end{figure}
\begin{figure}[!htbp]
\centering
\includegraphics[width=0.6\textwidth]{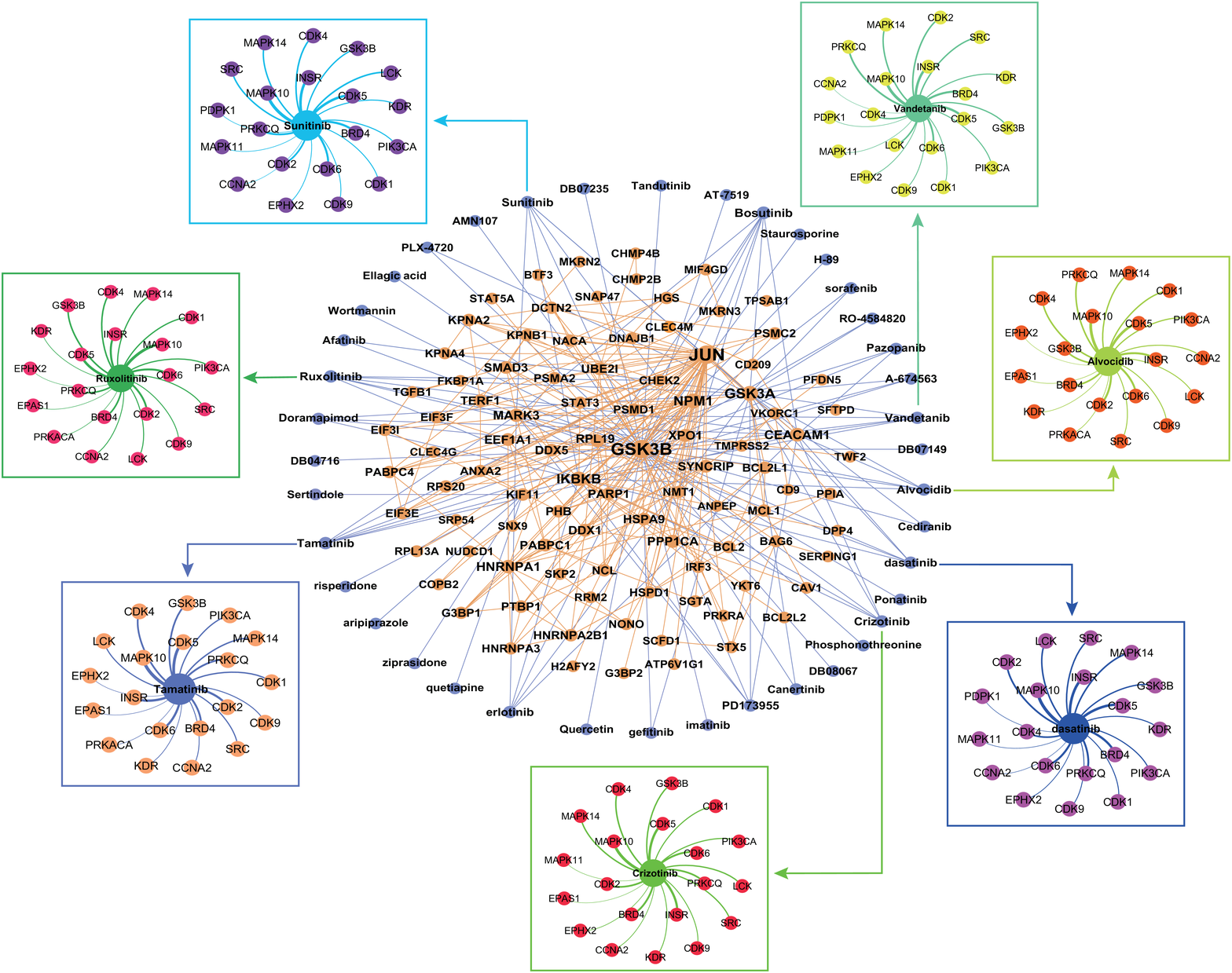}
\caption{Drug-protein associations networks. The center's network graph is composed of the known interactions of 119 HCoVs-host proteins in the PPI network (the orange edge) and is the drug-protein associations of the top 100 in the prediction ranking of the 119 HCoVs-host proteins (the blue edge).
Through a search of existing research literature, we found that 7 of the top 10 predicted drugs are related to coronavirus. We plotted the top 20 most relevant proteins predicted by these 7 drugs (see Supplementary data Table 4).}
\label{Figure.3}
\end{figure}

\subsection{Discovery of Drug Repurposing Candidates for COVID-19}
In addition to the assessment of the accuracy of the prediction results, we also analyze the drug-protein associations of the top 100 in the prediction ranking of the 119 HCoVs-host proteins (see Supplementary data Table 2) and draw a network graph (Figure.3). We find that Ruxolitinib and Imatinib in the clinical trial stage appeared in the network graph of Figure.3. To find drugs that  may be relevant in addition to clinical trials, we have researched in the literature on the prediction of the top 50 drugs (see Table Supplementary data Table 3), and find that 36 drugs have been mentioned in the past studies that may have a role in the treatment of coronavirus-related diseases. These drugs are likely to play a role in COVID-19. Tamatinib, Sunitinib, Ruxolitinib, Alvocidib, Crizotinib, Dasatinib, Vandetanib among the top 10 drugs have also been found related to coronavirus through searches in related literature. We plot the most relevant top 20 proteins of  these drugs in the network diagram. The most relevant top 20 proteins prediction results of all top 50 drugs can be found in Supplementary data Table 4. It also shows that it is effective to find drugs that may act on HCoVs-related targets through a comprehensive heterogeneous network.

\section{CONCLUSIONS AND DISCUSSION}
\label{others}

In this study, we used a deep learning model to systematically recognize drugs that may act on HCoVs-related proteins and promote the discovery and research of drug reuse in COVID-19. We used existing HCoVs-related host proteins and proteins directly related to these host proteins in the PPI  network to predict drugs that can be reused. We constructed a comprehensive heterogeneous network that connects drugs, HCoVs-related proteins, and diseases. We found the method showed high  performance.\\
At present, our research still has some limitations. In this study, we have calculated the drug reuse based on  the HCoVs-related host proteins described in the existing  literature. The collected virus-host interaction may not be complete, and  its quality may be affected by many factors. Nonetheless, our method has taken into account as much as possible the existing interactions, the expression of target proteins, the chemical structure of drugs, and similarities. In the future, we can use the calculation method to obtain  more useful and reliable information, and further systematically prediction interaction between the drug and the  target.\\

\subsubsection*{Acknowledgments}
This work was supported by the national key R\&D program of China (2017YFE0130600); the National Natural Science Foundation of China (Grant Nos.61772441,61872309, 61922020, 61425002, 61872007); Project of marine economic innovation and development in Xiamen(No. 16PFW034SF02); Natural Science Foundation of Fujian Province (No. 2017J01099). This paper is recommended by the 5th Computational Bioinformatics Conference.

\bibliography{ref}

\begin{thebibliography}{7}
\providecommand{\natexlab}[1]{#1}
\providecommand{\url}[1]{\texttt{#1}}
\expandafter\ifx\csname urlstyle\endcsname\relax
  \providecommand{\doi}[1]{doi: #1}\else
  \providecommand{\doi}{doi: \begingroup \urlstyle{rm}\Url}\fi

\bibitem[Bullinaria \& Levy(2007)Bullinaria and Levy]{bullinaria2007extracting}
John~A Bullinaria and Joseph~P Levy.
\newblock Extracting semantic representations from word co-occurrence
  statistics: A computational study.
\newblock \emph{Behavior research methods}, 39\penalty0 (3):\penalty0 510--526,
  2007.

\bibitem[Gysi et~al.(2020)Gysi, Do~Valle, Zitnik, Ameli, Gan, Varol, Sanchez,
  Baron, Ghiassian, Loscalzo, et~al.]{gysi2020network}
Deisy~Morselli Gysi, {\'I}talo Do~Valle, Marinka Zitnik, Asher Ameli, Xiao Gan,
  Onur Varol, Helia Sanchez, Rebecca~Marlene Baron, Dina Ghiassian, Joseph
  Loscalzo, et~al.
\newblock Network medicine framework for identifying drug repurposing
  opportunities for covid-19.
\newblock \emph{ArXiv}, 2020.

\bibitem[Hsieh et~al.(2015)Hsieh, Natarajan, and Dhillon]{hsieh2015pu}
Cho-Jui Hsieh, Nagarajan Natarajan, and Inderjit Dhillon.
\newblock Pu learning for matrix completion.
\newblock In \emph{International Conference on Machine Learning}, pp.\
  2445--2453. PMLR, 2015.

\bibitem[Niwattanakul et~al.(2013)Niwattanakul, Singthongchai, Naenudorn, and
  Wanapu]{niwattanakul2013using}
Suphakit Niwattanakul, Jatsada Singthongchai, Ekkachai Naenudorn, and
  Supachanun Wanapu.
\newblock Using of jaccard coefficient for keywords similarity.
\newblock In \emph{Proceedings of the international multiconference of
  engineers and computer scientists}, volume~1, pp.\  380--384, 2013.

\bibitem[Paules et~al.(2020)Paules, Marston, and Fauci]{paules2020coronavirus}
Catharine~I Paules, Hilary~D Marston, and Anthony~S Fauci.
\newblock Coronavirus infections—more than just the common cold.
\newblock \emph{Jama}, 323\penalty0 (8):\penalty0 707--708, 2020.

\bibitem[Zeng et~al.(2020)Zeng, Zhu, Lu, Liu, Huang, Zhou, Fang, Huang, Guo,
  Li, et~al.]{zeng2020target}
Xiangxiang Zeng, Siyi Zhu, Weiqiang Lu, Zehui Liu, Jin Huang, Yadi Zhou,
  Jiansong Fang, Yin Huang, Huimin Guo, Lang Li, et~al.
\newblock Target identification among known drugs by deep learning from
  heterogeneous networks.
\newblock \emph{Chemical Science}, 11\penalty0 (7):\penalty0 1775--1797, 2020.

\bibitem[Zhou et~al.(2020)Zhou, Hou, Shen, Huang, Martin, and
  Cheng]{zhou2020network}
Yadi Zhou, Yuan Hou, Jiayu Shen, Yin Huang, William Martin, and Feixiong Cheng.
\newblock Network-based drug repurposing for novel coronavirus
  2019-ncov/sars-cov-2.
\newblock \emph{Cell discovery}, 6\penalty0 (1):\penalty0 1--18, 2020.

\end{thebibliography}
\bibliographystyle{iclr2021_conference}

\end{document}